\pdfoutput=1

\documentclass[11pt]{article}

\usepackage[preprint]{acl}

\usepackage{times}
\usepackage{latexsym}

\usepackage[T1]{fontenc}

\usepackage[utf8]{inputenc}

\usepackage{microtype}

\usepackage{inconsolata}

\usepackage{graphicx}

\usepackage[utf8]{inputenc} 
\usepackage[T1]{fontenc}    
\usepackage{hyperref}       
\usepackage{url}            
\usepackage{booktabs}       
\usepackage{amsfonts}       
\usepackage{nicefrac}       
\usepackage{microtype}      
\usepackage{xcolor}         

\usepackage{bbding}
\definecolor{green}{HTML}{009B55}
\usepackage{microtype}
\usepackage{graphicx}
\usepackage{subfigure}
\usepackage{threeparttable}
\usepackage{booktabs} 
\usepackage{xspace}
\usepackage[ruled,vlined]{algorithm2e}
\usepackage{macros}
\usepackage{wrapfig}
\usepackage{lipsum}
\usepackage{amsfonts}       
\usepackage{nicefrac}       
\usepackage{xcolor, colortbl}         
\usepackage{multicol}
\usepackage{multirow}
\usepackage{enumitem}

\usepackage{listings}
\usepackage{pythonhighlight}

\usepackage{titletoc}
\usepackage{mathrsfs}
\usepackage{upquote}
\usepackage[linewidth=0.5pt,leftmargin=1em,rightmargin=1em,innertopmargin=5pt,
    innerbottommargin=5pt,
    innerrightmargin=5pt,
    innerleftmargin=5pt]{mdframed}
\usepackage[toc,page,header]{appendix}

\usepackage{times}
\usepackage{latexsym}
\usepackage{tabularx}
\usepackage{multirow}
\usepackage{booktabs} 
\usepackage{enumitem}
\usepackage{amsmath}

\usepackage{inconsolata}
\usepackage{graphicx}

\usepackage[utf8]{inputenc} 
\usepackage[T1]{fontenc}    
\usepackage{hyperref}       
\usepackage{url}            
\usepackage{booktabs}       
\usepackage{amsfonts}       
\usepackage{nicefrac}       
\usepackage{microtype}      
\usepackage{xcolor}         
\usepackage{multicol}
\usepackage{transparent}
\usepackage{array}
\usepackage{bm}

\definecolor{nblue}{cmyk}{0.95,0.0,0.2,0.2}

\usepackage{xspace}
\usepackage{cleveref}
\usepackage{colortbl}
\usepackage{mdframed}
\usepackage{listings}
\definecolor{green}{HTML}{009B55}

\usepackage{listings}
\usepackage{pythonhighlight}
\usepackage{fancyvrb}
\RecustomVerbatimCommand{\VerbatimInput}{VerbatimInput}{fontsize=\footnotesize,
 frame=single,  
 framesep=0.5em, 
 labelposition=topline,
}

\usepackage{tcolorbox}
\tcbuselibrary{listings}
\newcommand{\ours}{\textsc{MSQA}\xspace}

\title{\ours: Benchmarking LLMs on Graduate-Level Materials \\ Science Reasoning and Knowledge}

\author{Jerry Junyang Cheung$^1$\thanks{These authors contributed equally to this work.}, Shiyao Shen$^{1*}$, Yuchen Zhuang$^1$, Yinghao Li$^1$,\\
\textbf{Rampi Ramprasad$^2$, Chao Zhang$^1$}\\
$^1$College of Computing, $^2$School of Materials Science and Engineering\\
  Georgia Institute of Technology, Atlanta, USA \\
  \texttt{\{jzhang3027,sshen94,yczhuang,yinghaoli\}@gatech.edu} \\
  \texttt{rampi.ramprasad@mse.gatech.edu, chaozhang@gatech.edu}}

\begin{document}
\maketitle
\begin{abstract}
Despite recent advances in large language models (LLMs) for materials science, there is a lack of benchmarks for evaluating their domain-specific knowledge and complex reasoning abilities. 
To bridge this gap,
we introduce \ours{}, a comprehensive evaluation benchmark of 1,757 graduate-level materials science questions in two formats: detailed explanatory responses and binary True/False assessments. 
\ours{} distinctively challenges LLMs by requiring both precise factual knowledge and multi-step reasoning across seven materials science sub-fields, such as structure-property relationships, synthesis processes, and computational modeling.
Through experiments with 10 state-of-the-art LLMs, we identify significant gaps in current LLM performance. While API-based proprietary LLMs achieve up to 84.5\% accuracy, open-source (OSS) LLMs peak around 60.5\%, and domain-specific LLMs often underperform significantly due to overfitting and distributional shifts. 
\ours{} represents the first benchmark to jointly evaluate the factual and reasoning capabilities of LLMs crucial for LLMs in advanced materials science.

\end{abstract}

\section{Introduction}
\label{sec:intro}

\begin{figure}[t]
  \includegraphics[width=0.98\linewidth]{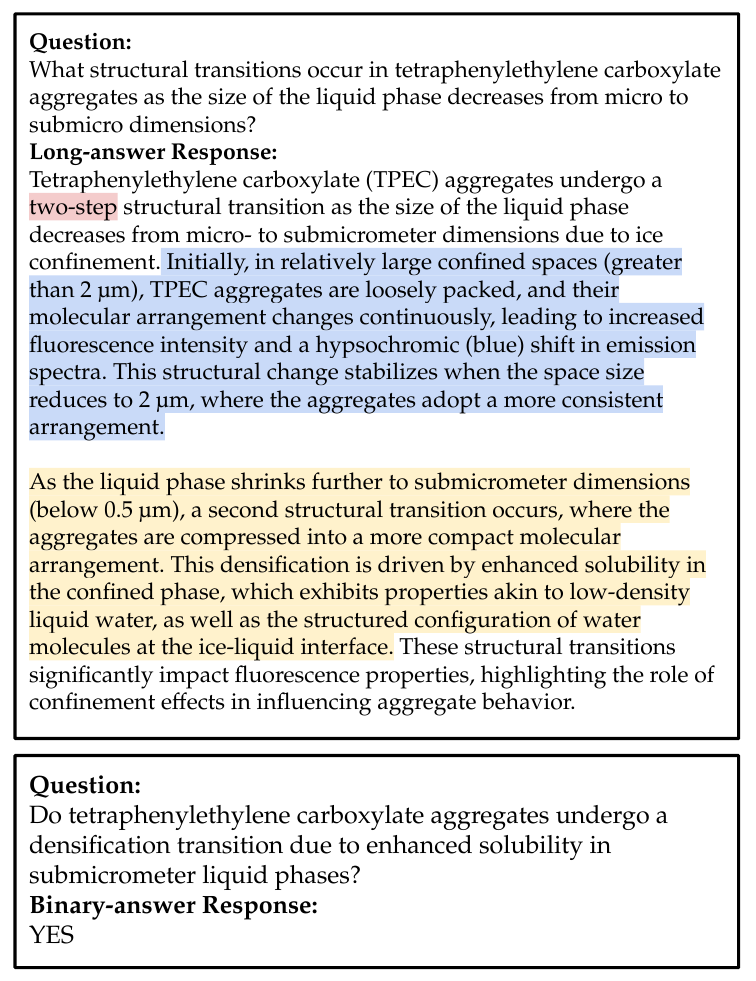}
    \caption{Example question-answer pairs from \ours{}. The long-answer response (top) demonstrates multi-step reasoning with explanations, while the binary-answer response (bottom) shows a concise yes/no determination on a related concept. 
    }
  \label{fig:example}
\end{figure}

Recent advances in large language models (LLMs) hold considerable promise for accelerating materials research by automating complex analysis~\cite{miret2024llms}, prediction~\cite{rubungo2024llm4mat}, and design tasks \cite{jablonka202314}. However, progress in applying LLMs to materials science has been constrained by the scarcity of high-quality domain-specific annotated datasets \cite{miret2024llms} and robust evaluation benchmarks \cite{zaki2023mascqa}.

Existing datasets for materials science primarily focus on conventional natural language processing (NLP) tasks, such as named entity recognition (NER) \cite{weston2019named, shetty2023general}, classification \cite{venugopal2021looking, gupta2022matscibert}, and relation extraction (RE) \cite{cheung2023polyie, song2023matsci}. 
Although valuable, these datasets do not sufficiently assess models' materials science knowledge grounded reasoning and question-answering abilities.
Recent efforts \cite{zaki2023mascqa} have introduced benchmarks featuring questions derived from graduate admission exams; however, the short-answer format (\eg, multiple choice) limits the assessment of complex reasoning and in-depth explanatory capabilities essential for real-world applications in materials science, such as multistep synthesis planning and detailed property evaluations.

To bridge this critical gap, we introduce \ours{}, a graduate-level materials science benchmark specifically crafted to evaluate both factual knowledge and complex reasoning capabilities of LLMs. 
\ours includes two complementary evaluation modes: \emph{long-answer} and \emph{binary-answer} (Figure \ref{fig:example}). 
The long-answer questions demand detailed, multi-step explanations spanning seven challenging subfields, including structure-property relationships, polymer synthesis, and computational material modeling. 
In contrast, the binary-answer questions offer balanced 
True/False queries that require LLMs to assess complex domain-specific judgments on material properties, applications, and technical claims.
Together, \ours{} comprehensively test the depth of factual understanding and advanced reasoning skills.

To ensure high-quality and domain-grounded questions and answers, we employ advanced LLMs, including \texttt{gpt-4o} \cite{hurst2024gpt}, \texttt{gemini-2.0-pro} \cite{team2023gemini}, and \texttt{Deepseek-v3} \cite{deepseekai2025deepseekr1incentivizingreasoningcapability}, guided by expert-curated materials science literature. 
The dataset generation process incorporates rigorous three-stage quality assurance: (1) regular expression-based filtering, (2) LLM-driven refinement, and (3) expert annotation. 

In our experiments, we systematically benchmark seven leading open-source and black-box LLMs alongside three domain-specific fine-tuned models. 
Our results reveal that commercial black-box LLMs consistently outperform open-source alternatives, achieving accuracy as high as 84.5\%. 
Incorporating retrieved contextual data notably enhances model performance, showing retrieval augmentation as a crucial adaptation strategy. 
Conversely, domain-specific fine-tuned models underperform relative to general-purpose models, likely due to distribution shifts and overfitting, underscoring critical limitations in current domain-adaptation approaches.
We summarize our main contributions as follows:


\begin{itemize}[leftmargin=0.6cm]
    \item We present \ours, one of the first materials science benchmarks explicitly designed to rigorously test complex reasoning and explanatory abilities of LLMs beyond factual knowledge; 
    \item We provide a thorough empirical evaluation of leading general-purpose and domain-specific LLMs. We conduct detailed analyses of challenging scenarios to catalyze the development of more reasoning-intensive, domain-adapted LLMs for the materials science community. 
    \item We open-source our curated dataset and benchmark code to foster community-driven innovation towards LLM-driven advanced material science discovery: \url{https://github.com/jerry3027/MSQA}.
\end{itemize}

\section{Related Works}
\label{sec:related}

\noindent \textbf{Materials Science Datasets for LLMs.}
Prior research in materials science NLP primarily targets structured extraction tasks such as NER and RE. 
Expert-curated datasets, such as \citet{wang-etal-2021-chemner} and \citet{weston2019named}, focus explicitly on identifying and extracting material names, properties, and their interrelations. 
Additional specialized datasets emphasize tasks like property prediction; for instance, \citet{friedrich-etal-2020-sofc} annotated a corpus of scholarly articles related to solid oxide fuel cells, tagging entities such as \texttt{materials}, \texttt{values}, and \texttt{devices}, while \citet{panapitiya-etal-2021-extracting} provided annotations on chemical entities (\texttt{CHEM}), numerical values (\texttt{VALUE}), and measurement units (\texttt{UNIT}) from studies on soluble materials.
More recent datasets aimed at evaluating LLMs include question-answer (QA) pairs to test domain knowledge.
\citet{zaki2023mascqa} created a dataset with 650 questions derived from graduate-level admissions exams in India, while \citet{song2023matsci} aggregated multiple previously published datasets into a meta-dataset. 
However, they primarily utilize short-answer formats such as multiple-choice or numerical values, which inadequately capture the nuanced reasoning and explanatory capabilities required in real-world materials science applications.


\noindent \textbf{Synthetic Data Generation for Benchmarks.} 
LLMs increasingly serve as tools for creating evaluation benchmarks, especially when manual curation requires domain expertise or is prohibitively expensive.
For example, SciBench \cite{wang2024scibenchevaluatingcollegelevelscientific} and BioGPTQA \cite{Sarwal2023.12.19.572483} employ LLM-generated content subsequently validated through expert reviews and structured filtering mechanisms. 
Other benchmarks, including MT-Bench \cite{zheng2023judgingllmasajudgemtbenchchatbot} and HELM \cite{liang2023holisticevaluationlanguagemodels}, similarly rely on synthetically generated data to evaluate model performance across diverse tasks. 
Synthetic data generation markedly decreases annotation expenses; however, it is important to ensure the validity and reliability of generated benchmarks. Researchers employ quality control measures such as expert validation, statistical filtering, and alignment with reference materials to minimize factual inaccuracies and reasoning shortcuts.

\section{\ours: A Graduate-Level Materials Science QA Dataset}
\begin{figure*}[]
    \centering
    \includegraphics[width=\textwidth]{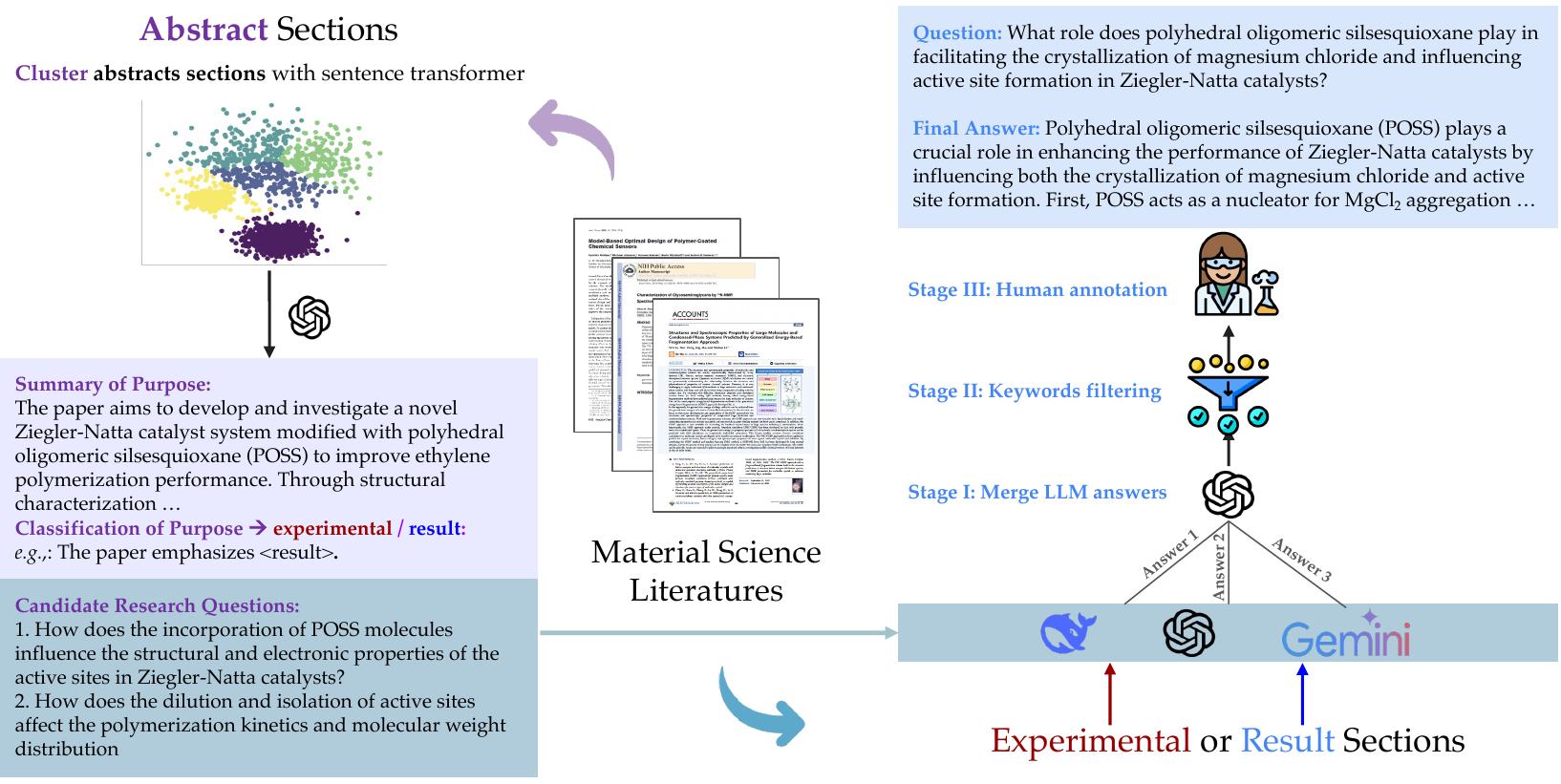}
    \caption[]{Overview of data generation and quality evaluation in \ours.}
    \label{fig:pipline_diagram}
\end{figure*}

\label{sec:method}


We present \ours, a comprehensive materials science benchmark for evaluating LLMs (Figure~\ref{fig:pipline_diagram}). 
We begin by describing our scientific literature collection process in \cref{subsec:data-collection}. We then detail the procedures for generating long-answer questions and their corresponding answers in \cref{subsec:question-generation} and \cref{subsec:answer-generation}, respectively. Next, we outline our three-stage quality assurance process in \cref{subsec:quality-assurance}. We then describe the generation process for binary-answer questions in \cref{subsec:binary-question}. Finally, we summarize key dataset statistics for \ours{} in \cref{subsec:dataset-statistics}.


\subsection{Data Collection}
\label{subsec:data-collection}

Previous research has highlighted the inherent limitations of LLMs in effectively addressing questions grounded in specialized materials science knowledge \cite{miret2024llms, zaki2023mascqa, wang2024evaluating}. To overcome this limitation, we curate a comprehensive collection of materials science literature to serve as a context for generating high-quality QA pairs.
Specifically, we source 3,000 diverse articles from the extensive dataset compiled by \citet{shetty2023general}, encompassing 2.4 million publications from seven primary materials science publishers spanning from 2000 to 2021. We employ sentence transformers \cite{reimers-2019-sentence-bert} to generate embeddings for abstracts from 50,000 randomly selected articles to ensure diversity and representativeness. These embeddings are subsequently clustered using K-means to achieve the highest silhouette score \cite{9260048}, resulting in 10 distinct clusters, each with 5000 papers on average. From each cluster, we randomly sample 300 articles, culminating in our final set of 3,000 context articles.



Instead of using PDF documents, we collected papers in XML format and parsed them using a publicly available \texttt{chemistry-paper-parser} tool.\footnote{\url{https://github.com/Yinghao-Li/ChemistryHTMLPaperParser}} This allows accurate information extraction from these papers by preserving the integrity of mathematical formulas and chemical representations.

\subsection{Candidate Question Generation}\label{subsec:question-generation}
To ensure that the generated questions accurately reflect the complexity and depth of real-world materials science research, we utilize the abstracts of selected articles, as they succinctly encapsulate research objectives, methodologies, and significant findings.
Following \citet{zhong2024actionie}, we first prompt \texttt{gpt-4o} to generate concise summaries highlighting key findings from each abstract. This initial summarization step reduces the influence of specialized scholarly language, thereby facilitating more precise and targeted question generation. 
Subsequently, each article is categorized based on its primary objective: introducing a new synthesis method (``method'') or presenting novel experimental observations (``result''). 
Guided by this categorization, \texttt{gpt-4o} is then prompted to formulate candidate questions specifically aligned with the focus identified in the article.



\subsection{Question Selection and Candidate Answer Generation}
\label{subsec:answer-generation}
Previous studies have shown that context relevance \cite{wang2024searching} and factual precision \cite{ram2023context} are crucial factors for response quality. 
To enhance alignment between questions and provided context, we employ a backward selection approach, where we use \texttt{gpt-4o} to select the candidate questions based on the provided context. 
Specifically, for articles categorized under ``methods'', we provide the \texttt{Experimental Method} sections containing detailed descriptions of research protocols and materials synthesis procedures. For articles categorized as ``results'', we supply the \texttt{Results} sections, which include comprehensive interpretations of experimental outcomes. 

Candidate answers were then generated using \texttt{gpt-4o}, \texttt{gemini-2.0-pro}, and \texttt{deepseek-v3}.
Initial assessments highlight that LLM-generated answers frequently included ambiguous references such as ``\texttt{the K0 Samples} and \texttt{the SCAs Units}'', diminishing clarity and self-contained informativeness. 
To address these issues, we refine the prompts by explicitly discouraging the use of definite articles, 
significantly enhancing answer clarity. 
This adjustment results in more precise and contextually anchored responses, explicitly referencing chemical entities (\eg, \texttt{hexamethyldisilane}, \texttt{copper phthalocyanine}) and specific methodologies (\eg, \texttt{CryoTEM Imaging}, \texttt{dynamic light scattering}) from the original studies. 

\subsection{Three-Stage Quality Assurance}
\label{subsec:quality-assurance}

Prior research \cite{wang2022self, huang2022large} indicates that self-consistency among LLMs significantly enhances answer accuracy and coherence. Extending this idea, \citet{li2025llms} applies self-consistency to handle open-ended tasks effectively. Inspired by these findings, we adopt a multi-model self-consistency approach to enhance answer quality. Specifically, candidate answers produced by \texttt{gpt-4o} \cite{hurst2024gpt}, \texttt{gemini-2.0-pro} \cite{team2023gemini}, and \texttt{deepseek-v3} \cite{deepseekai2025deepseekr1incentivizingreasoningcapability} (\cref{subsec:answer-generation}) are aggregated using \texttt{gpt-4o}. This aggregation explicitly accounts for inter-model agreement, leveraging consensus among multiple sophisticated models. 

To ensure question clarity and relevance, we first employ automated filtering methods, using regular expressions and keyword matching to remove ambiguous and overly context-dependent questions. Subsequently, materials science domain experts manually review the remaining questions to exclude unclear, incorrect, or irrelevant queries.


We recruit two materials science PhD students to rigorously evaluate the quality of a representative subset of the generated answers. Specifically, we randomly select 50 question-answer pairs for assessment. Each evaluator independently applied their expert domain knowledge to assess whether the provided answers are: (1) factually correct, (2) directly relevant and precisely addressed the questions, and (3) logically coherent. 
The evaluators show that 92.86\% of answers fulfilled all three quality criteria. 

\subsection{Binary Question Generation}
\label{subsec:binary-question}
Given the computational expense associated with evaluating detailed long-form answers using advanced LLMs such as \texttt{gpt-4o}, we develop a set of 1,757 binary True/False questions derived from previously generated question-answer pairs. 
Initial efforts to directly convert existing pairs into binary format reveal several issues: (1) questions frequently included overly detailed clues revealing the correct answer; (2) LLM-generated questions demonstrated a pronounced bias towards ``True'' responses; and (3) the generated questions lacked complexity, often omitting nuanced reasoning involving approximations or comparative thresholds (\eg, ``exact value'' versus ``around'' a value).

To tackle these issues, we instruct \texttt{gpt-4o} to (1) explicitly generate questions with a predefined True or False label, yielding 878 ``True'' and 879 ``False'' labeled questions; (2) favor approximations (\eg, ``around'', ``more or less than'') for rigid details that are not essential; and (3) avoid unnecessary detail without adding challenge.
Detailed prompts are included in \cref{app:prompt}. 

\subsection{Dataset Statistics}
\label{subsec:dataset-statistics}

\begin{table}[t]\small
\centering
    \begin{tabular}{l  c}
        \toprule
        \textbf{Question \& Answer Pairs} & \textbf{Counts} \\
        \midrule
        Long-answer Q\&A pairs & 1,757 \\
        Binary-answer Q\&A pairs & 1,757 \\ 
        \hspace{2mm}- \# w/ ``True'' label & 878 \\
        \hspace{2mm}- \# w/ ``False'' label & 879 \\
        Avg. question length (in words) & 19 \\
        Avg. long answer length (in words) & 150 \\
        \bottomrule
    \end{tabular}
\caption{Data statistics of \ours.} 
\label{table:statistics}
\end{table}

Table \ref{table:statistics} summarizes the basic statistics of \ours{}.

\noindent \textbf{Question Types.} 
We manually annotate a subset of questions to identify and categorize various question types pertinent to general materials science tasks. Using \texttt{gpt-4o} \cite{hurst2024gpt}, we then classify the remaining questions into these predefined categories. For questions with ambiguous types, \texttt{gpt-4o} is prompted to label them as ``Other'', which are subsequently manually reviewed and categorized by domain experts. In total, we identify seven distinct question categories (Table~\ref{table: 1}) that comprehensively assess LLM capabilities within the materials science domain.

\begin{table}[t]\small
\centering
    \begin{tabular}{l  c}
        \toprule
        \textbf{Question Types} & \textbf{Counts} \\
        \midrule
        Structure-property relationships & 818 \\
        Synthesis and processing & 257 \\ 
        Computational & 216 \\
        Material analysis techniques & 187 \\
        Material modeling & 125 \\
        Failure analysis and degradation & 93 \\
        Material properties & 61 \\
        \bottomrule
    \end{tabular}
\caption{Question composition in \ours.}
\label{table: 1}
\end{table}

\begin{figure}[]
    \centering
    \includegraphics[scale=0.48]{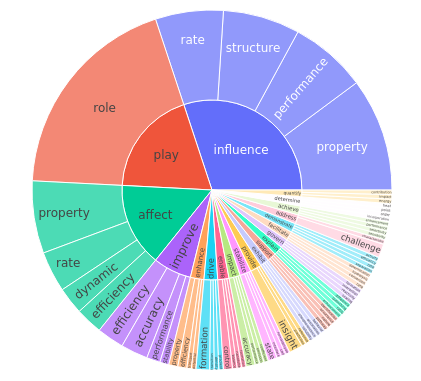}
    \caption[]{The top 20 most common root verbs (inner circle) and their top 4 direct noun objects (outer circle) in the generated instructions.}
    \label{fig:verb-noun-pie}
\end{figure}

\noindent \textbf{Question Semantics.} 
To examine question semantics, we analyze their verb-noun structures following the methodology of \citet{wang2023selfinstructaligninglanguagemodels}. We employ the Berkeley Neural Parser\footnote{\url{https://parser.kitaev.io/}} to parse each question, extracting the primary verb (closest to the root) and its direct noun object. The most frequent root verbs, along with their associated direct noun objects, are visualized in Figure \ref{fig:verb-noun-pie}. This analysis highlights the broad topical coverage and complex conceptual nature of questions in \ours{}, particularly emphasizing relationships between material designs, their properties, and relevant experimental methodologies.

\begin{figure*}
    \centering
    \includegraphics[width=\textwidth]{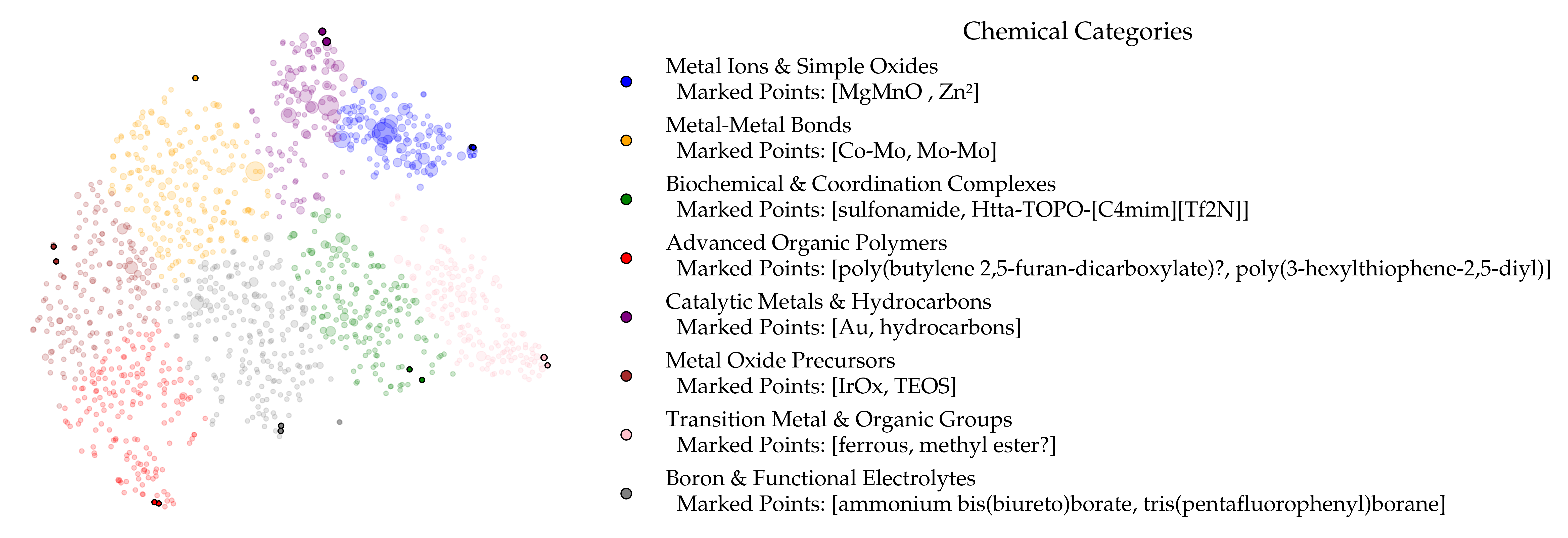}
    \caption[]{T-SNE visualization of material and chemical mentions from long-form answers, embedded using MatSciBERT. Legend shows manually labeled categories with example compounds.}
    \label{fig:tsneFig}
\end{figure*}

\noindent \textbf{Answer Semantics.}
To further explore the diversity of the dataset, we analyze the semantic content of the long-form answers by identifying references to specific materials and chemical compounds using ChemDataExtractor \cite{swain2016chemdataextractor}. 
Extracted chemical entities are embedded using MatSciBERT \cite{gupta_matscibert_2022}, a specialized language model trained on extensive materials science literature.
These embeddings are then visualized through t-SNE clustering, presented in Figure \ref{fig:tsneFig}. 
Upon examining the resulting clusters, we confirm that the answers encompass diverse material categories, including \texttt{Polymers \& Copolymers} and \texttt{Inorganic Complexes}, illustrating the comprehensive topical diversity inherent in our dataset.

\section{Experiments}
\label{sec:exp}

\subsection{Experimental Setup}

\noindent \textbf{Tasks.} 
We evaluate the performance of open-source and domain-specific LLMs on both the long-answer and binary-answer variants of the \ours dataset using two distinct prompting strategies: direct generation and retrieval-augmented generation. In the direct-generation scenario, models are presented only with the question, without supplementary context. In contrast, in the retrieval-augmented setting, we first build a contextual database by segmenting the Methods and Results sections described in \cref{subsec:question-generation} into separate paragraphs. We then use BM25 \cite{robertson2009probabilistic} to retrieve the top five most relevant paragraphs, which serve as additional context provided to the model.
For binary-answer questions, we also investigate if chain-of-thought \cite{wei2022chain} improves model performance.
Black-box LLMs are evaluated exclusively under the direct generation setting. Evaluation of the long-answer responses is conducted through GPT-4o acting as an LLM judge, assessing responses as either ``correct'' or ``mostly correct'', both categories counted as correct in our metrics. For binary-answer evaluations, accuracy is determined by exact keyword matching for responses containing either ``YES'' or ``NO''. Performance results are uniformly reported as accuracy percentages across all experiments.

\noindent \textbf{Evaluated LLM models.} We comprehensively evaluate several LLMs on \ours{}. For black-box models, our analysis includes \texttt{Claude-3.7-Sonnet} \cite{anthropicClaudeSonnet}, \texttt{Gemini-2.0-Flash}, and \texttt{Grok-3} \cite{xGrokBeta}. Additionally, we assess the performance of prominent open-source models, specifically \texttt{Llama-3-8B} \cite{grattafiori2024llama}, \texttt{Phi-4-mini} \cite{abdin2024phi}, \texttt{Qwen-2.5-7B} \cite{yang2024qwen2}, and \texttt{Deepseek-R1-distilled-Llama-3}.
Furthermore, we benchmark several domain-specific LLMs specialized for materials science. These include \texttt{Honeybee} \cite{song2023honeybee}, an LLM fine-tuned iteratively on synthesized materials science data; \texttt{Mol-Instructions-Molecule} \cite{fang2023mol}, a model fine-tuned explicitly on chemical reaction and molecular design datasets tailored for small molecules; and \texttt{Llasmol} \cite{yu2024llasmol}, a specialized chemistry domain LLM instruction-tuned across 14 chemistry-specific tasks utilizing a dataset exceeding three million samples.

\noindent \textbf{LLM-as-Judge.} 
Prior studies have consistently validated the effectiveness of employing LLMs for pairwise comparisons across diverse applications \cite{qin2023large, liu2024aligning, liusie2023llm}. Moreover, \citet{zeng2023evaluating} demonstrate that incorporating rule-based or self-generated evaluation criteria from LLMs further enhances the accuracy and reliability of these assessments. Given that our dataset involves comprehensive long-answer responses comprising detailed explanations of synthesis processes and materials modeling, we leverage \texttt{gpt-4o} as an evaluator with a gold-standard reference answer, a model-generated inference response, and a structured evaluation rubric. \texttt{gpt-4o} then evaluates the inference response as ``correct'', ``mostly correct'', or ``incorrect''. We further verify the validity of \texttt{gpt-4o}'s judgment by comparing it against human expert evaluations detailed in \cref{subsec:llm-as-judge}.

\begin{table*}[t]
    \centering
    \fontsize{7.5}{9.5}\selectfont\setlength{\tabcolsep}{0.3em}
    \resizebox{\linewidth}{!}{
    \begin{tabular}{lccccc}
        \toprule
        \textbf{Tasks ($\rightarrow$)} & \multicolumn{2}{c}{\textbf{Long-answer}} & \multicolumn{3}{c}{\textbf{Binary-answer}}\\
        \cmidrule(lr){2-3} \cmidrule(lr){4-6}
        \textbf{Baselines ($\downarrow$)}& DP & RAG & DP & COT & RAG \\
        \midrule
        \multicolumn{6}{l}{\textit{Open-source LLMs}}\\
        \midrule
        Llama-3-8B~\cite{grattafiori2024llama} & 39.39 (16/676/1065) & 85.20 (330/1167/260) & 63.97 & 57.37 & 73.71 \\
        Phi-4-mini~\cite{abdin2024phi} & 46.39 (15/800/942) & 51.28 (207/694/856) & 68.24 & 60.39 & 64.43 \\
        Qwen-2.5-7B~\cite{yang2024qwen2} & 51.28 (41/860/856) & 87.48 (504/1033/220) & 72.34 & 69.89 & 83.84 \\
        Deepseek-R1-distilled-Llama3~\cite{deepseekai2025deepseekr1incentivizingreasoningcapability} & 60.50 (37/1026/694) & 85.71 (362/1144/251) & 52.74 & 51.91 & 65.40 \\
        \midrule
        \multicolumn{6}{l}{\textit{Domain-specific LLMs}}\\
        \midrule
        Honeybee~\cite{song2023honeybee} & 19.53 (0/343/1413) & 2.73 (3/45/1708) & 8.82 & 0.23 & 0.68 \\
        Mol-Instructions-Molecule~\cite{fang2023mol} & 0.23 (0/4/1753) & 6.66 (16/101/1640) & 22.82 & 0.11 & 11.84 \\
        Llasmol~\cite{yu2024llasmol} & 4.84 (0/85/1672) & 6.26 (12/98/1647) & 28.34 & 5.41 & 29.82 \\
        \bottomrule
    \end{tabular}
    }
    \caption{Main results of open-source and domain-specific LLMs on \ours. Long-answer results are presented as ``accuracy'' in \% (Correct/Mostly Correct/Incorrect). Binary-answer results are presented as ``accuracy'' in \%. Notations are consistent across tables. DP refers to ``direct prompting''. COT refers to ``direct prompting with chain of thoughts''.}
    \label{tab:main-results}
\end{table*}

\subsection{Main Results}
Tables \ref{tab:main-results} and \ref{tab:black-box-llms} present the results for open-source, domain-specific, and black-box LLMs on \ours, respectively.

\noindent \textbf{Long-Answer Questions.} 
From Table \ref{tab:main-results}, we highlight several key observations: (1) among the evaluated open-source models, \texttt{Deepseek-R1-distilled-Llama3} \cite{deepseekai2025deepseekr1incentivizingreasoningcapability} achieves the highest accuracy (60.50\%), outperforming \texttt{Qwen-2.5-7B} \cite{yang2024qwen2} (51.28\%) and \texttt{Phi-4-mini} \cite{abdin2024phi} (46.39\%). The superior performance of \texttt{Deepseek-R1-distilled-Llama3} may be attributed to the model's inherent self-correction nature in its thought process, allowing it to review and refine its outputs;
(2) retrieval-augmented generation (RAG) notably improves performance for \texttt{Llama-3} and \texttt{Qwen-2.5-7B}. However, \texttt{Phi-4-mini} exhibits only a marginal improvement from 46.39\% to 51.28\%, likely due to limited exposure to long-context and retrieval-augmented training data in its alignment corpus. This highlights the crucial role of post-training model alignment; and
(3) domain-specific LLMs surprisingly underperform compared to general-purpose models in producing coherent long-form answers. This underperformance is potentially due to distributional shifts between their specialized finetuning datasets and our more general domain-focused dataset, alongside evident overfitting. 
For instance, \texttt{LlamaSmol} \cite{yu2024llasmol} model frequently outputs chemical names encapsulated within \texttt{<SMILE>} tags, reflecting such training limitations.


\noindent \textbf{Binary-Answer Questions.} 
As shown in Table \ref{tab:main-results}, binary-answer questions present considerable difficulty, with two out of four general-purpose models (\texttt{Llama-3-8B} and \texttt{Deepseek-R1-distilled-Llama3}) performing only slightly above random guessing levels at 63.97\% and 52.74\%, respectively. 
Interestingly, chain-of-thought \cite{wei2022chain} decreases performance for all open-source and domain-specific LLMs. Our analysis reveals that this decline is due to LLMs generating factually incorrect intermediate steps, likely stemming from their limited materials science knowledge, as confirmed by \citet{wang2024evaluating}.
Domain-specific models again perform worse than general-purpose models, with the best domain-specific model, \texttt{Llasmol}, achieving only 28.34\% accuracy. We attribute this performance gap to two causes: 1) the domain-specific models likely overfitted on the finetuning data, decreasing their ability to output "True" and "False" answers; 2) the distribution shift between our dataset and their finetuning corpus. 

\noindent \textbf{Black-Box LLMs.} 
Results presented in Table \ref{tab:black-box-llms} clearly demonstrate that black-box LLMs substantially outperform open-source models in answering long-form questions. Specifically, Grok-3 \cite{xGrokBeta} achieves an impressive accuracy of 84.46\% without supplementary contextual data. However, performance on binary-answer tasks remains comparable between black-box and open-source models, with slight improvement after chain-of-thought \cite{wei2022chain} prompting. 



\begin{table}\small
\centering
    \begin{tabular}{lccc}
        \toprule
        \textbf{Model} & \textbf{Long-ans} & \textbf{Binary-ans} & \textbf{Binary-cot}\\
        \midrule
        Claude & 66.35 (136/840/495) & 68.58 & 70.18\\
        Grok & 84.84 (363/885/223) & 65.05 & 71.37 \\
        Gemini & 77.63 (254/888/329)& 72.85 & 71.54\\
        \bottomrule
    \end{tabular}
    \caption{Experimental results of black-box LLMs on \ours with direct prompting. 
    }
    \label{tab:black-box-llms}
\end{table}

\subsection{LLM as Judge}
\label{subsec:llm-as-judge}
\begin{figure}
    \centering
    \includegraphics[width=\linewidth]{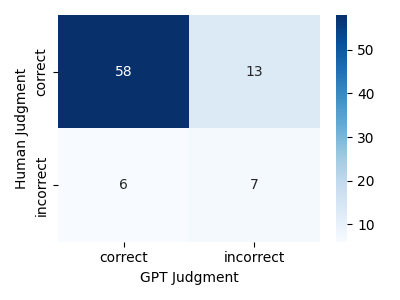}
    \caption{Confusion matrix between human and GPT-4o judgments.}
    \label{fig:confusion-matrix-gpt-alignment}
\end{figure}

We investigate the reliability of using an LLM as a judge by comparing its annotation decisions directly against those made by human annotators. Considering human annotations as the gold standard, we quantified agreement using a confusion matrix, as illustrated in Figure \ref{fig:confusion-matrix-gpt-alignment}. Results indicate that \texttt{gpt-4o}'s evaluations align with human judgments in approximately 77.38\% of cases, demonstrating particularly high consistency for answers rated as ``correct''.

\begin{table}\small
\centering
{ %
    \begin{tabular}{lc}
        \toprule
        \textbf{Model} & \textbf{Accuracy (\%)} \\
        \midrule
        Llama-3-8B & 77.18\\
        Phi-4-mini & 92.26\\
        Qwen-2.5-7B & 95.11\\
        Deepseek-R1-distilled-Llama3& 96.07\\
        \bottomrule
    \end{tabular}
    }
    \caption{Results on long-answer questions with GPT-4o-mini as judge.}
    \label{tab:4o-mini-as-judge}
\end{table}

\noindent \textbf{GPT-4o-mini as Judge.} 
Due to the computational expense associated with GPT-4o, we explore the viability of using GPT-4o-mini as an alternative judge for assessing long-answer responses generated by open-source LLMs (Table \ref{tab:4o-mini-as-judge}). 
We observe that GPT-4o-mini demonstrates several distinct biases compared to GPT-4o. Notably, GPT-4o-mini exhibits a pronounced \emph{verbosity bias} \cite{saito2023verbosity}, often incorrectly rating longer responses as accurate, particularly evident in evaluations involving Deepseek-distilled-r1-llama3. Additionally, GPT-4o-mini frequently \emph{accepts vague or irrelevant explanations} as correct. For instance, when asked "{How does the presence of an SH-OC hydrogen bond influence the geometry and stability of the global minimum conformer of methyl 3-mercaptopropionate?}", GPT-4o-mini deemed a general explanation related to hydrogen bonding sufficient. Moreover, GPT-4o-mini \emph{fails to consistently identify scientific inaccuracies}. 
Specifically, it incorrectly classified an inference answer that labeled Li$_4$SnS$_4$ as an \emph{anode material} rather than correctly as a \emph{solid electrolyte}, a mistake accurately detected by GPT-4o.


\begin{table} \small
\centering
    \begin{tabular}{lc}
        \toprule
        \textbf{Question Type} & \textbf{Accuracy (\%)} \\
        \midrule
        structure-property relationships & 49.27\\
        synthesis and processing & 40.86\\
        computational & 49.07\\
        material analysis techniques & 45.45\\
        material modeling & 45.6\\
        failure analysis and degradation & 39.78\\
        material properties & 36.07\\
        \bottomrule
    \end{tabular}
    \caption{Results of Qwen-2.5-7B on each question type.}
    \label{tab:question-type}
\end{table}

\subsection{Question Type Difficulties}
We conducted a case study to evaluate the performance of Qwen-2.5-7B \cite{yang2024qwen2} across various categories of materials science questions, as summarized in Table \ref{tab:question-type}. The model exhibited the lowest accuracy (36.07\%) on questions related to \emph{material properties}, likely attributable to the lack of symbolic understanding of numerical values. This finding aligns with prior research by \citet{miret2024llms}, which similarly underscores the difficulties LLMs encounter with materials science numerical tasks. The \emph{failure analysis and degradation} category emerged as the second most challenging, probably due to the sparse representation of these topics within its pre-training dataset. Conversely, Qwen-2.5-7B demonstrated superior performance on questions involving \emph{structure-property relationships}, indicating stronger foundational knowledge likely driven by the broader availability of material structure data in textbooks and journal articles.

\subsection{Error Analysis}
We conduct a detailed manual analysis of inference errors made by LLMs on the \ours dataset and identify several recurrent failure patterns: (1) LLMs explicitly recognize their own limitations in domain-specific knowledge and thus fail to provide complete answers; (2) LLMs deliver partially accurate responses, neglecting critical sub-questions or necessary qualifying statements; and (3) LLMs generate scientifically incorrect or misleading facts due to hallucination. We include specific examples illustrating each of these error types in \cref{app:error-analysis}.





\section{Conclusion}
\label{sec:conclusion}

In this study, we introduce \ours, a comprehensive benchmark explicitly designed to assess LLMs on complex, domain-specific reasoning and explanatory capabilities in materials science. 
Comprising 1,757 rigorously crafted long-answer and binary-answer questions, \ours addresses a significant gap in current evaluation resources by simulating realistic scientific inquiry scenarios. 
Our extensive evaluation of ten advanced LLMs highlights substantial performance gaps, particularly revealing limitations in accurately generating coherent, nuanced responses to complex materials science queries. 
\ours serves as a robust platform for benchmarking and advancing the development of specialized LLMs tailored to the demanding context of materials science research.

\section*{Limitations}
We rely on GPT-4 as a judge for evaluating long-answer responses. However, this approach introduces evaluation costs for future researchers wishing to replicate or extend our work. Due to constraints on annotation resources, we performed manual annotations on a randomly sampled subset of the questions. This process, while necessary, may result in the inclusion of a small number of low-quality question pairs due to the inherent variability in LLM-generated content. Third, our computational limitations restricted us from evaluating open-source LLMs exceeding 8 billion parameters. We acknowledge that this may limit the generalizability of our findings. We encourage future research to overcome these limitations by assessing larger-scale LLMs.

\bibliography{ref}

\appendix
\section{License}
This dataset is licensed under the MIT License. Future works are free to use, modify, and distribute this dataset in accordance with the terms of the MIT License. 

\section{Prompt Details}
\label{app:prompt}
\textbf{Prompt for generating candidate questions:}
\begin{Verbatim}
Here is the "abstract" of a materials science
paper. Please complete the following tasks:

1. Summarize the purpose of the paper in clear
and concise terms. 
2. Classify the purpose as emphasizing 
"<method>" or "<result>".
3. Identify research questions relevant to the
abstract's themes and materials science 
interests.

"Abstract": {paper[key_abstract]}
\end{Verbatim}
\textbf{Prompt for generating candidate answers:}
\begin{Verbatim}
I will provide the purpose of a materials 
science paper, related research questions, and 
a detailed section of the paper.

Your tasks:
    1. Select the Most Relevant Question:
Choose the research question that is most 
specific, clearly phrased, and directly 
related to the provided section.
    2. Refine the Question: Modify the selected 
question to ensure it is:
    - Grounded on information from the 
provided section, but answerable even 
without using the provided section.
    - Standalone and unambiguous. Do not 
use definite articles when referring to 
compounds.
    - Clearly phrased for precision.
    3. Generate a Direct Answer: Provide a 
concise and well-structured response that:
    - Directly answers the question.
    - Is based on the provided section but 
remains meaningful out of context.
    - Avoids vague references such as 
"this study" or "this paragraph."
    - Clearly conveys the information 
without requiring the reader to see the 
original section.

Present the output as a JSON shown below:
{{
    "question": "A clear and specific 
question.",
    "answer": "A concise and relevant answer 
that remains meaningful without additional 
context."
}}

    Input data:
    - "Purpose and related questions": 
{llm_curate_abstract}
    - "Detailed section": 
{paper[key_detail]}
\end{Verbatim}

\textbf{Prompt for merging candidate answers:}
\begin{Verbatim}
### Instructions:
1. Review the above solutions.
2. Generate an improved and refined solution by 
aggregating the strengths from the provided 
solutions. Enclose the solution within 
<SOLUTION> and </SOLUTION> tag.
3. Provide a brief explanation of your 
reasoning.
4. Ensure your answer is clear, concise, and 
structured logically.

### Here is the problem:
"question": {question},

### Reference Solutions:
Solution 1: {answer1}

Solution 2: {answer2}

Solution 3: {answer3}
\end{Verbatim}

\textbf{Prompt for LLM-as-Judge:}
\begin{Verbatim}
Your task is to evaluate the accuracy of 
LLM-generated answers to materials science 
questions by comparing them to expert-validated 
"gold" answers.

For each evaluation, you will receive:
	- A materials science question
	- A gold answer, based on authoritative
domain knowledge
	- An LLM-generated inference answer, 
which you must assess

Your goal is to evaluate how well the inference
answer aligns with the gold answer in terms of 
factual accuracy, conceptual completeness, and 
relevance.

Use the following evaluation rubric:
    - Correct: The inference answer fully
captures all essential concepts from the gold 
answer, with no significant omissions or 
factual errors.
    - Mostly Correct: The inference answer 
conveys the main idea or correct conclusion, 
even if minor details are missing or slight 
inaccuracies are present. Additional 
non-conflicting information is acceptable.
    - Incorrect: The inference answer 
demonstrates substantial misunderstanding, 
includes major factual errors, or omits core 
concepts present in the gold answer.

Provide a short justification for your rating, 
highlighting key similarities or discrepancies 
between the inference and gold answers. Output 
your response in the following JSON format:
{{
    "reasoning": "A concise explanation 
supporting your judgment.",
    "judgment": 
"correct|mostly correct|incorrect"
}}
\end{Verbatim}

\section{Keywords for candidate question filtering}
\label{app:keywords}
We use the following keywords to filter vague candidate questions that make direct reference to the context:

\noindent["this", "article", "paper", "study", "abstract", "experimen", "studied", "investigat", "discuss", "describ", "example", "document", "section", "the finding","the sample"].

\section{Error Analysis}
\label{app:error-analysis}
\subsection{Cases where LLM admits that it does not have knowledge about the question}

\textbf{Question:}
How does destruction of Mn4CaO5 cluster affect conformational flexibility of CP43 lumenal domain?

\noindent\textbf{LLM Answer:}
    
\noindent \emph{I'm not aware of any information regarding how the destruction of the Mn4CaO5 cluster affects the conformational flexibility of the CP43 lumenal domain.} The Mn4CaO5 cluster is a critical component of the photosynthetic water-oxidizing complex (WOC) in photosystem II, a complex protein in photosynthetic organisms. The CP43 lumenal domain is part of this WOC.

The Mn4CaO5 cluster is essential for the water-oxidation reaction during photosynthesis, where light energy is used to split water into oxygen, protons, and electrons. The destruction of the Mn4CaO5 cluster can be caused by various factors, such as oxidative stress, mutations, or chemical modifications, and it can disrupt the water-oxidation process.

However, I am unable to provide any detailed information regarding how this affects the conformational flexibility of the CP43 lumenal domain. To answer this question, I would need to look into more specific research related to this topic.

\subsection{Case where LLM only partially address the question (Unaddressed part is bolded)}

\textbf{Question:}
How does the band structure of methylammonium lead iodide influence its intrinsic semiconductor behavior \textbf{despite the presence of defects and impurities?}

\noindent\textbf{LLM Answer:}
    
\noindent The band structure of methylammonium lead iodide (MAPbI$_3$), a widely studied perovskite material, is characterized by a direct band gap at the Brillouin zone center ($\tau$-point), with the valence band maximum primarily derived from the iodine 5p orbitals and the conduction band minimum from the lead 6p orbitals. This direct band gap enables efficient light absorption and emission, making MAPbI$_3$ an excellent absorber for photovoltaic applications. The relatively small band gap (~1.55 eV) allows absorption across the visible spectrum, while strong spin-orbit coupling due to lead further influences the band structure, enhancing carrier mobility. The presence of organic cations (methylammonium) has minimal direct impact on the electronic states near the band edges but contributes to structural stability, allowing for flexible tuning of the material's optoelectronic properties.

\subsection{Case where LLM introduces incorrect scientific facts (Incorrect facts are bolded)}

\textbf{Question:}
How do variations in polysarcosine block length affect protein adsorption and surface passivation in block copolymers?

\noindent\textbf{LLM Answer:}
    
\noindent Polysarcosine (PSA) is a water-soluble, biocompatible, and biodegradable polymer that has been studied extensively in the context of biomedical applications, including drug delivery, tissue engineering, and surface modification. Block copolymers comprising polysarcosine and another polymer (often referred to as PSA-based block copolymers) are of particular interest due to their ability to self-assemble into nanostructures with unique properties.

Variations in polysarcosine block length in PSA-based block copolymers can significantly affect protein adsorption and surface passivation. Here's a general overview of how different block lengths can impact these properties:

1.  **Protein Adsorption:**   

*   Shorter PSA blocks: When the PSA block is relatively short (e.g., 4-10 monomers), the block copolymer \textbf{tends to adsorb more proteins} from the surrounding environment due to the relatively hydrophobic nature of the core block (often a hydrophobic polymer like poly(L-lysine) or poly(ethylene oxide)). This is because the hydrophobic core can interact more strongly with hydrophobic parts of proteins, leading to enhanced protein adsorption.

*   Longer PSA blocks: Conversely, when the PSA block is longer (e.g., 20-50 monomers), the block copolymer tends to exhibit reduced protein adsorption. This is because the increased hydrophilicity of the PSA block dominates the surface, creating a more hydrophilic interface that repels proteins.

2.  **Surface Passivation:**

*   Shorter PSA blocks: Short PSA blocks can lead to poor surface passivation due to the underlying hydrophobic core being more accessible to the environment, resulting in a higher likelihood of protein adsorption and biofouling.

*   Longer PSA blocks: Longer PSA blocks are \textbf{more effective at surface passivation}. They create a more stable and hydrophilic surface that reduces protein adsorption, minimizing the risk of biofouling and promoting a more stable interface with the surrounding environment.

In summary, the length of the polysarcosine block in PSA-based block copolymers plays a significant role in determining protein adsorption and surface passivation properties. Shorter PSA blocks tend to facilitate protein adsorption, while longer PSA blocks enhance surface passivation.

\section{Annotation Details}
\label{app:annotation-details}

\subsection{Annotation Guideline}
\begin{figure*}[]
    \centering
    \fbox{\includegraphics[width=\textwidth]{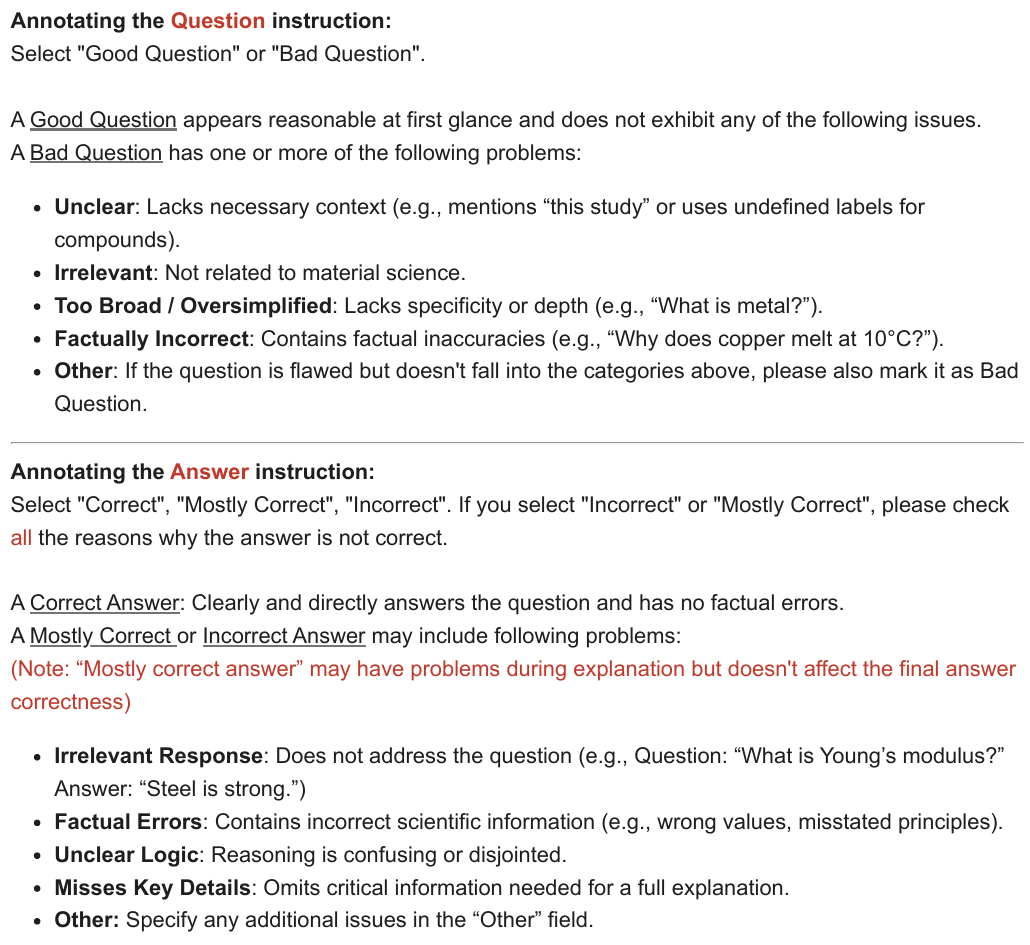}}
    \caption{Annotation Survey Guideline}
    \label{fig:Annotation-guidelines}
\end{figure*}

The annotation guideline is shown in Figure \ref{fig:Annotation-guidelines}. We detail the criteria for the annotation of QA pairs in \ours.

\subsection{Annotation Format}
\begin{figure*}[]
    \centering
    \fbox{\includegraphics[width=\textwidth]{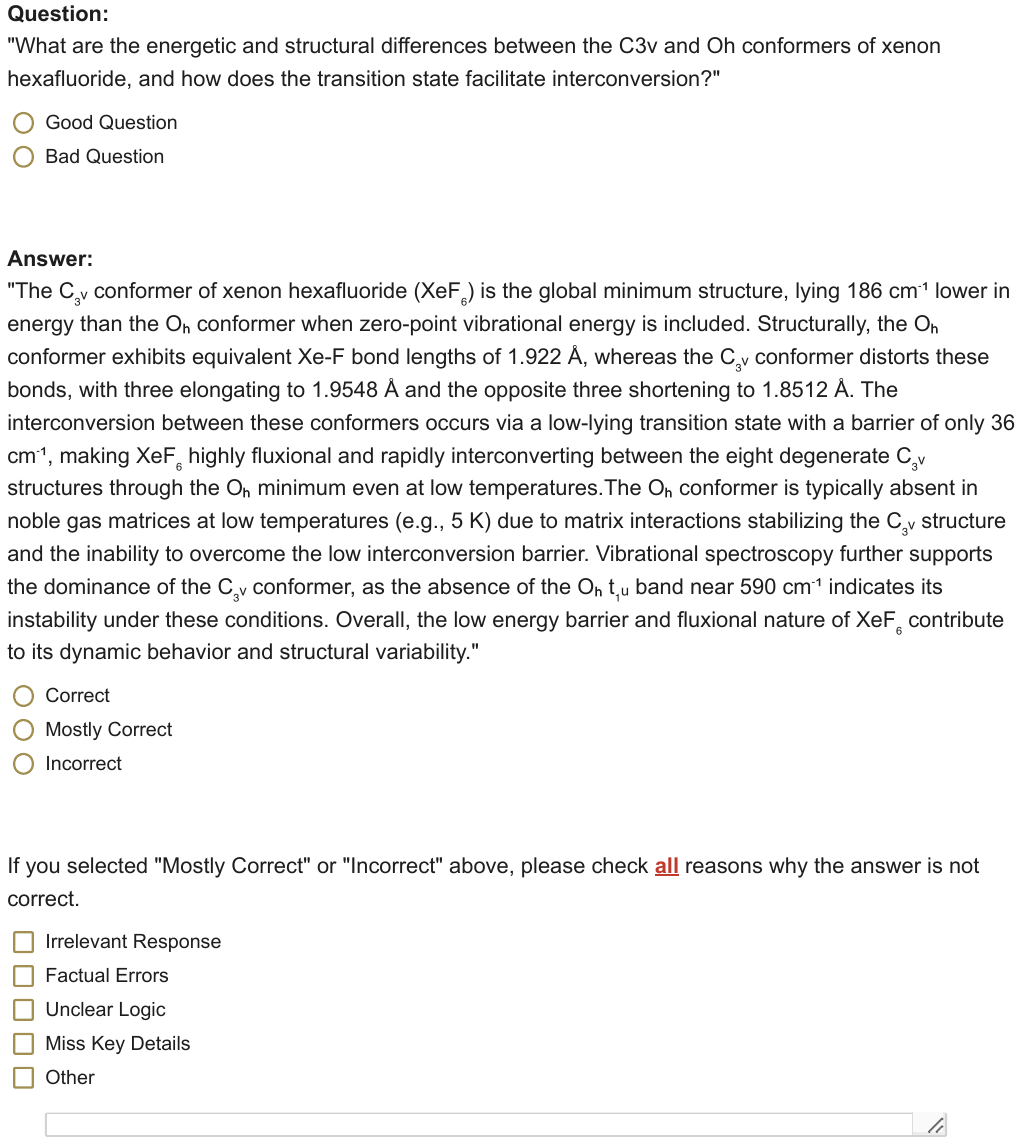}}
    \caption{Annotation Survey Format}
    \label{fig:Annotation-Format}
\end{figure*}

Screenshot of the annotation survey is shown in Figure \ref{fig:Annotation-Format}. For question quality, we ask the annotator to annotate whether the question is a ``Good Question'' or ``Bad Question''. For answer quality, we ask them to identify whether the answers are ``Correct'', ``Mostly Correct'', or ``Incorrect'', matching the judgment choices of LLM judge. If the annotator selected ``Mostly Correct'' or ``Incorrect'', they are required to specify their reasoning.

\subsection{Annotation Agreement}
\label{app:annotation-agreement}
\begin{figure}
    \centering
    \includegraphics[width=\linewidth]{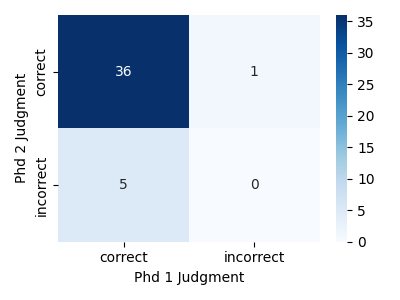}
    \caption{Confusion matrix between two material science expert annotators.}
    \label{fig:gold-confusion-matrix}
\end{figure}

Confusion matrix for annotation agreement between two PhD students on the "gold" generated data for \ours is shown in Figure \ref{fig:gold-confusion-matrix}.

\end{document}